\documentclass{article}

\usepackage{arxiv}

\usepackage[utf8]{inputenc} % allow utf-8 input
\usepackage[T1]{fontenc}    % use 8-bit T1 fonts
\usepackage{amsmath}        % for \text{} in math mode
\usepackage{url}            % simple URL typesetting
\usepackage{booktabs}       % professional-quality tables
\usepackage{amsfonts}       % blackboard math symbols
\usepackage{nicefrac}       % compact symbols for 1/2, etc.
\usepackage{microtype}      % microtypography
\usepackage{lipsum}
\usepackage{graphicx}
\graphicspath{ {./images/} }
\usepackage{tikz}
\usepackage[ruled,vlined]{algorithm2e}
\usepackage{times}

\usepackage{amssymb}
\usepackage{graphicx}
\usepackage{multirow}
\usepackage{float}
\usepackage{authblk}
\usepackage{hyperref}

\title{Fine-Tuning Video Transformers for Word-Level Bangla Sign Language: A Comparative Analysis for Classification Tasks}

\author{
    Jubayer Ahmed Bhuiyan Shawon$^{1*}$, Hasan Mahmud$^1$, Kamrul Hasan$^1$\\
    $^1$\textit{Systems and Software Lab (SSL), Department of CSE, Islamic University of Technology (IUT),}\\
    \textit{OIC, Board Bazar, Gazipur-1704, Bangladesh.}\\
    \textit{*Corresponding author:} \href{mailto:juabyerahmed@iut-dhaka.edu}{juabyerahmed@iut-dhaka.edu}\\
    \textit{Contributing authors:} 
    \href{mailto:hasan@iut-dhaka.edu}{hasan@iut-dhaka.edu}; 
    \href{mailto:hasank@iut-dhaka.edu}{hasank@iut-dhaka.edu}
}

\begin{document}
\maketitle
\begin{abstract}
Sign Language Recognition (SLR) involves the automatic identification and classification of sign gestures from images or video, converting them into text or speech to improve accessibility for the hearing-impaired community. In Bangladesh, Bangla Sign Language (BdSL) serves as the primary mode of communication for many individuals with hearing impairments. This study fine-tunes state-of-the-art video transformer architectures—VideoMAE, ViViT, and TimeSformer—on BdSLW60 \cite{rubaiyeat2025bdslw60}, a small-scale BdSL dataset with 60 frequent signs. We standardized the videos to 30 FPS, resulting in 9,307 user trial clips. To evaluate scalability and robustness, the models were also fine-tuned on BdSLW401 \cite{rubaiyeat2025bdslw401transformerbasedwordlevelbangla}, a large-scale dataset with 401 sign classes. Additionally, we benchmark performance against public datasets, including LSA64 and WLASL. Data augmentation techniques such as random cropping, horizontal flipping, and short-side scaling were applied to improve model robustness. To ensure balanced evaluation across folds during model selection, we employed 10-fold stratified cross-validation on the training set, while signer-independent evaluation was carried out using held-out test data from unseen users U4 and U8. Results show that video transformer models significantly outperform traditional machine learning and deep learning approaches. Performance is influenced by factors such as dataset size, video quality, frame distribution, frame rate, and model architecture. Among the models, the VideoMAE variant (MCG-NJU/videomae-base-finetuned-kinetics) achieved the highest accuracies—95.5\% on the frame rate corrected BdSLW60 dataset and 81.04\% on the front-facing signs of BdSLW401—demonstrating strong potential for scalable and accurate BdSL recognition.
\end{abstract}

% keywords can be removed

\section{Introduction}
More than 430 million people worldwide, including approximately 34 million children, experience some form of hearing loss—constituting about 5\% of the global population. Alarmingly, this number is projected to double by 2050, underscoring the urgent need for scalable and effective communication solutions for the deaf and hard-of-hearing community \cite{WHO2021}. Sign languages, which rely on intricate combinations of hand gestures, movements, postures, and facial expressions, serve as the primary mode of communication for many individuals with hearing impairments \cite{article1}. However, communication remains a significant challenge, as most hearing individuals lack fluency in sign language. This communication barrier is further exacerbated by the scarcity, high cost, and limited accessibility of professional sign language interpreters, thereby impeding the social inclusion and daily interaction of deaf individuals \cite{deepasl}.

Sign Language Recognition (SLR) aims to bridge this gap by leveraging computer vision and machine learning to automatically interpret sign language gestures \cite{ZHANG20242399}. SLR approaches are typically divided into two categories: isolated recognition, which focuses on identifying individual signs or fingerspelling frames, and continuous recognition, which interprets temporal sequences of signs to form phrases or sentences \cite{9079505,article2}. While continuous SLR must contend with ambiguous sign boundaries and temporal segmentation, isolated SLR operates at the gloss level, where each video contains exactly one sign.

A variety of isolated sign language datasets—such as AUTSL, LSA64, WLASL, BosphorusSign22k, and LSM—have propelled research forward in languages like Turkish, Argentinian and American Sign Language \cite{Mercanoglu_Sincan_2022, Vazquez-Enriquez_2021_CVPR, s22135043, Laines_2023_CVPR, ronchetti2023lsa64argentiniansignlanguage}. In contrast, studies on isolated Bangla Sign Language (BdSL) remain limited by data scarcity and resource constraints. The subtle intra‑class variations and fine-grained hand movements in BdSL further complicate classification tasks, making high‑accuracy recognition challenging \cite{zhao2023bestbertpretrainingsign}.

Early efforts in SLR relied on traditional machine learning methods with hand‑crafted features, which often struggled with scalability and robustness. The advent of deep learning (DL)—particularly convolutional and recurrent neural networks—has substantially improved visual recognition performance across many domains \cite{dlservey, ZHANG20242399}, including isolated SLR \cite{sharma2021asl, liang20183d}. However, naive attempts to train attention-based DL architectures from scratch on BdSL data have yielded suboptimal performance, underscoring the need for more effective model adaptation strategies \cite{rubaiyeat2025bdslw60}.
%-------- not cited -------------------------------------

Recently, transformer‑based architectures—notably video transformers \cite{novopoltsev2023finetuningsignlanguagerecognition, videomae_, du2022full} and detection transformers— \cite{detr} have demonstrated strong capabilities in modelling spatiotemporal dependencies for word‑level sign recognition. Transfer learning \cite{Chen_2022_CVPR, 4270344, DAS2023118914, s20185151, mocialov2020transferlearningbritishsign, app132111625}, wherein pre‑trained models are fine‑tuned on domain‑specific data, has emerged as a powerful tool for boosting accuracy in data‑scarce scenarios. Key considerations in transfer learning include the selection of which layers to transfer and whether to freeze or fine‑tune them \cite{finetuning}.

While prior studies have focused primarily on recognizing Bangla sign letters and numerals \cite{shawon, N+A, bornonet}, limited work has addressed word-level BdSL recognition using modern deep learning techniques such as EfficientNet-B3, attention-based transformers, and BiLSTM \cite{empath, bdsl_opa, rubaiyeat2025bdslw60}. To the best of our knowledge, no prior research has applied video transformers to word-level BdSL recognition.

In this work, we explore the potential of pre-trained video transformers—such as VideoMAE, ViViT, and TimeSformer—for isolated BdSL recognition. These models are fine-tuned on the BdSLW60 dataset and their generalization capabilities are assessed on the larger BdSLW401 dataset. While earlier efforts in BdSL have predominantly focused on static fingerspelling or character-level recognition, our approach targets dynamic, word-level recognition using spatiotemporal modeling. Furthermore, we extend our evaluation across other sign language datasets to investigate the robustness of these models under varying frame rates, class distributions, and video quality, thereby addressing key challenges in low-resource sign language recognition.
In short, the key contributions of our work are as follows:

\begin{enumerate}
\item We examine accuracy fluctuations resulting from FPS correction and improve performance by introducing variations in uniformly chosen frames.
\item We provide the first large-scale benchmark of the transformer-based Video Models (VideoMAE, ViViT, TimeSformer) fine-tuned on isolated BdSL datasets and also demonstrate generalization across other sign language datasets (LSA64, WLASL100, WLASL2000)
\item We analyze the impact of frame imbalance, FPS (25, 30, 60) in small (BdSLW60, LSA64, WLASL100) to large-scale datasets (BdSLW401, WLASL2000), per-class sample size, model architecture, and video quality on the performance of video transformers.
\end{enumerate}
We organise this work as follows: Section II provides
a survey of relevant literature and analyses contemporary
methodologies and their deficiencies.
Section III outlines the proposed architecture, dataset preparation, an overview of the video transformers, configurations, and the fine-tuning process.
Section IV outlines experimental results and performance
evaluation. Section V concludes the paper by delineating
prospective research avenues.

\section{Literature Review}
Sign language is a rich form of visual communication that encompasses both manual elements (hand movements, posture, position) and non-manual aspects (facial expressions, head gestures), assessed by traditional machine learning and deep learning methods \cite{madhiarasan2022comprehensivereviewsignlanguage}. Models such as support vector machines (SVM), hidden markov models (HMM), artificial neural networks (ANNs), and multilayer perceptrons (MLPs) have been applied with handcrafted feature extraction methods, including DCT, PCA, LDA, SURF, and SIFT, to improve classification accuracy \cite{lit_review, depth}.

  Several early efforts reported promising results. Al-Rousan et al. \cite{ALROUSAN2009990}, for instance, used HMM and DCT to classify 30 Arabic signs, achieving 94.2\% accuracy in signer-independent settings. Similarly, Fagiani et al. \cite{italiansign} applied HMM to 147 Italian signs, although the accuracy reached only 50\%, suggesting the need for more expressive models. Deep learning (DL) approaches emerged as a more powerful alternative, automating feature extraction and enabling end-to-end learning. A BiLSTM-based model with DeepLabv3+ hand segmentation achieved 89.5\% accuracy for 23 Arabic signs \cite{9079505}, while Fatmi et al. \cite{8666491} found that ANN and SVM outperformed HMM in American Sign Language (ASL) recognition.

More recently, hybrid DL models have improved word-level recognition. Masood et al. \cite{cnn+rnn} integrated inception-based CNNs with RNNs for real-time Argentine sign language detection. Similarly, ResNet50 combined with LSTM was used in \cite{cnn+lstm} for Persian sign videos, achieving accurate recognition across 100 signs. In another approach, spatial-temporal features from pretrained networks were fused, achieving 98.97\% accuracy on the Montalbano dataset \cite{alexnet+lstm}. CNN-transformer combinations have also shown promise: Shin et al. \cite{KSL} used such a hybrid to attain 88.8\% accuracy on a 77-class Korean Sign Language (KSL) dataset.

Research on Bangla Sign Language (BdSL) has also progressed. Raihan et al. \cite{SE+CNN} introduced channel-wise attention using squeeze-and-excitation blocks in a CNN model, reaching 99.86\% accuracy on the KU-BdSL alphabet dataset with a lightweight model optimized for mobile deployment. In another study, Begum et al. \cite{bornonet} utilized quantization on YOLOv4-Tiny with LSTM to achieve 99.12\% accuracy on the BdSL49 dataset. Other works have employed pose-based recognition using tools like OpenPose and Mediapipe. For example, \cite{openpose} used OpenPose for Flemish sign recognition, while \cite{mediapipe, rubaiyeat2025bdslw60, empath, mediapipe1} explored Mediapipe-based keypoints for dynamic body part tracking in Arabic and Bangla sign language recognition.

Attention mechanisms and transformer architectures have also made significant contributions. Rubayeat et al. \cite{rubaiyeat2025bdslw60} applied attention-based BiLSTMs with SVM to BdSLW60, achieving 75.1\% accuracy, and Hasan et al. \cite{empath} used an attention-based transformer for BdSL word-level recognition. These advances illustrate the growing utility of attention-based models, especially when combined with pose or spatiotemporal features. Knowledge transfer and transfer learning have further improved performance, as shown by \cite{knowledge_transfer}, who used MobileNetV2 with transfer learning to achieve 95.12\% accuracy on CSL-500 and a 2.2\% word error rate on CSL-continuous. Follow-up studies on BdSL also leveraged pretrained models like DenseNet201 and MobileNetV2 to boost recognition accuracy \cite{TL1, TL2}.

In parallel, video transformers have emerged as powerful tools for sign language recognition due to their ability to model complex temporal and spatial dependencies. For example, a study in \cite{videomae_} evaluated several video transformer models, including VideoMAE and SVT, on the WLASL2000 dataset—demonstrating the effectiveness of pretraining and fine-tuning in large-scale sign language recognition. Similarly, Detection Transformers (DETR) have been adapted to identify signs from RGB video inputs \cite{detr}. Beyond sign language, BERT has been combined with TimeSformer to improve the classification of short video clips \cite{timesformer}, and the ViViT model has been applied to detect mild cognitive impairment from video sequences, showing competitive performance \cite{mcvivit}.

Recognizing the limitations in existing BdSL research—particularly the underutilization of transformer-based video models—this work explores isolated BdSL word recognition using state-of-the-art video transformers. We fine-tune models pretrained on the Kinetics-400 action recognition dataset, which includes gestures and movements similar to isolated sign language actions. Our focus is on improving recognition performance from raw RGB videos using models such as VideoMAE, ViViT, and TimeSformer. Additionally, we analyze critical design factors such as frame distribution, frame rate (FPS), model architecture, and data quality that influence recognition outcomes.

\section{Methodology} 
\begin{figure}[t]
    \centering
    \includegraphics[width=0.8\columnwidth, height=0.4\textheight]{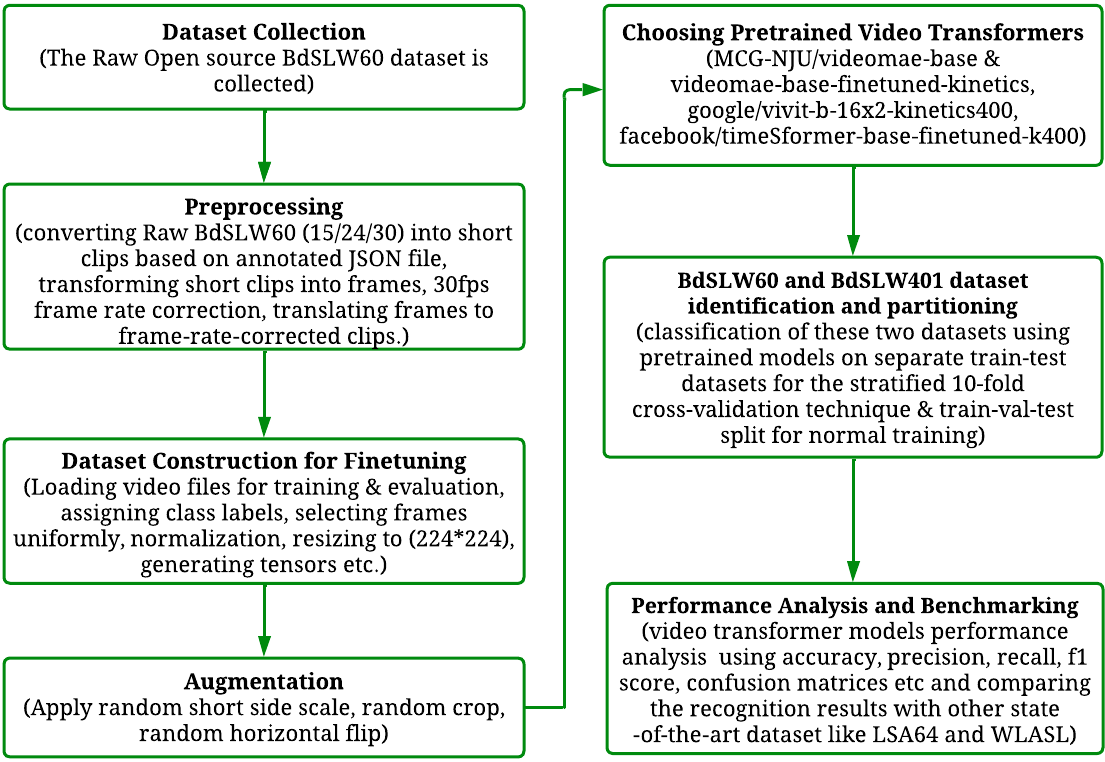}
    \caption{Architecture of Frame Rate-Corrected Dataset Construction, Recognition, and Benchmarking.}
    \label{fig:workflow}
\end{figure}

This study addresses the classification of isolated BdSL signs, with particular attention to the challenges posed by limited resources and the inherent complexity of sign language datasets. To tackle these issues, we fine-tune transformer-based video classification models to effectively capture temporal patterns in sequential data. In this work, three models are trained on BdSLW60, and to assess scalability, BdSLW401 is used—marking its first use as a benchmark. Their performance is rigorously evaluated and compared against results from other benchmark datasets, including WLASL and LSA64, to assess generalization and robustness. Figure~\ref{fig:workflow} presents the overall architecture of the proposed approach, illustrating the complete pipeline from dataset acquisition through preprocessing, model training, and ultimately, evaluation.

\subsection{Dataset preparation}
\begin{figure}[H] % frame distribution
    \centering
    \includegraphics [width=0.8\columnwidth, height=0.38\textheight]{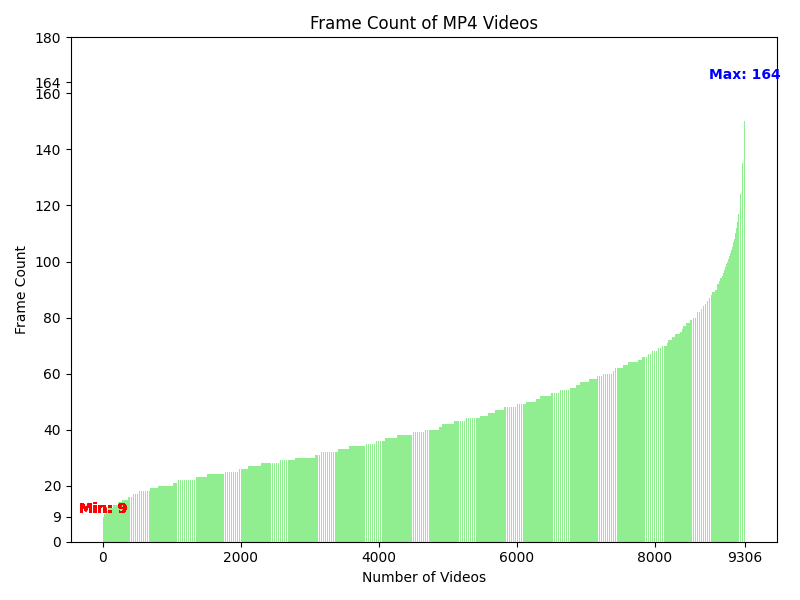}
    \caption{Frame Count vs Number of short clips of BdSLW60 dataset}
    \label{fig:frame_count_vs_video}
\end{figure}

This study utilizes the BdSLW60 dataset \cite{rubaiyeat2025bdslw60}, which requires careful preprocessing to ensure reliable training and evaluation. The dataset, sourced from Kaggle, comprises Bangla sign language videos recorded by 18 individuals. Each sample includes raw video footage and corresponding gloss annotations provided in JSON format. The videos were originally recorded at varying frame rates—15, 24, and 30 FPS—but were standardized to 30 FPS for consistency. The JSON annotations facilitated the extraction of individual frames, resulting in clip lengths ranging from 9 to 164 frames per gloss. As shown in Figure~\ref{fig:frame_count_vs_video}, most samples contain fewer than 130 frames. In addition to BdSLW60, we employed the BdSLW401, WLASL100, WLASL2000, and LSA64 datasets to comprehensively evaluate the performance and generalizability of our approach.

\subsection{Video Processing}
Determining the appropriate sample rate—defining the duration of each video clip used in training—requires a thorough understanding of the frame distribution across the dataset. This step is particularly critical for transformer-based models, which rely on fixed-length input sequences due to the patch embedding mechanism. The clip duration is calculated using the following equation: 
$\mathcal{C} = \frac{\mathcal{N} \times \mathcal{S}}{\mathcal{F}}$,  
where $\mathcal{C}$ is the clip duration, $\mathcal{N}$ is the number of frames to sample, $\mathcal{S}$ is the sample rate, and $\mathcal{F}$ is the frame rate (fps).

To mitigate the impact of class imbalance and ensure robust performance evaluation, we adopted a 10-fold stratified cross-validation procedure. Frame rate-corrected (FRC) samples were employed to partition the dataset into training, validation, and testing subsets for video processing tasks, or into training and testing sets for stratified folding.
The number of frames per clip was determined based on model-specific requirements and a predefined sampling rate. Clips with fewer frames than required were temporally padded to match the duration of the longest clip, ensuring input consistency. Additionally, all video frames were resized to 224×224 pixels and standardized to maintain uniformity and compatibility across model architectures.

\subsection{Augmentation for training}
During the data augmentation phase, we applied several techniques to improve dataset diversity and model generalization. These included random cropping, horizontal flipping, and short-side scaling. The isolated sign videos were subsequently trained in batches using three fine-tuned transformer-based models: VideoMAE \cite{tong2022videomaemaskedautoencodersdataefficient}, ViViT \cite{arnab2021vivitvideovisiontransformer}, and TimeSformer \cite{bertasius2021spacetimeattentionneedvideo}. To optimize frame-level feature extraction, we utilized image processors specifically designed for each pretrained model architecture. Model performance was continuously monitored on both validation and test sets throughout the training process. Upon completion, the best-performing model checkpoints were uploaded to the Hugging Face (HF) repository for public access and reproducibility.

\subsection{Model Configurations:}
We employed three transformer architectures for training and evaluation, with one model implemented in two distinct configurations, as summarized in Table~\ref{tab:model_comparison}. All models were pretrained on the Kinetics-400 human action recognition dataset \cite{kay2017kineticshumanactionvideo}, which enables the transfer of temporal and spatial pattern recognition capabilities applicable to isolated sign classification. Fine-tuning on the BdSLW60 and auxiliary datasets led to high accuracy on both test and validation splits, demonstrating the models' ability to effectively generalize to the target task.

\renewcommand{\arraystretch}{1.0} % Reduce row height
\setlength{\tabcolsep}{4pt} % Tighten column spacing
\begin{table}[ht]
    \centering
    \small % Smaller font size
    \begin{tabular}{lccc}
        \toprule
        \textbf{Model Name} & \textbf{MCG-NJU/videomae} & \textbf{google/vivit} & \textbf{facebook/TimeSformer} \\
        & \textbf{-base \& -finetuned-kinetics} & \textbf{-b-16x2-kinetics400} & \textbf{-base-finetuned-k400} \\
        \midrule
        \texttt{image\_size} & 224 & 224 & 224 \\
        \texttt{initializer\_range} & 0.02 & 0.02 & 0.02 \\
        \texttt{intermediate\_size} & 3072 & 3072 & 3072 \\
        \texttt{num\_attention\_heads} & 12 & 12 & 12 \\
        \texttt{num\_channels} & 3 & 3 & 3 \\
        \texttt{num\_frames} & 16 & 32 & 8 \\
        \texttt{num\_hidden\_layers} & 12 & 12 & 12 \\
        \texttt{patch\_size} & 16 & - & 16 \\
        \texttt{tubelet\_size} & 2 & [2, 16, 16] & 2 \\
        \texttt{hidden\_act} & gelu & gelu\_fast & gelu \\
        \texttt{decoder\_hidden\_size} & 384 & - & - \\
        \texttt{decoder\_intermediate\_size} & 1536 & - & - \\
        \texttt{decoder\_num\_attention\_heads} & 6 & - & - \\
        \texttt{decoder\_num\_hidden\_layers} & 4 & - & - \\
        \texttt{use\_mean\_pooling} & true & - & - \\
        \textbf{Trainable Parameters} & 94.2M \& 86.5M & 86M & 121M \\
        \bottomrule
    \end{tabular}
    \vspace{5mm}  
    \caption{Comparison of different video transformer models and their architecture details.}
    \label{tab:model_comparison}
\end{table}

\subsubsection{Video Mask Auto Encoder—VideoMAE architecture}
\begin{figure}[ht]
    \centering
    \includegraphics[width=1\columnwidth, height=0.2\textheight]{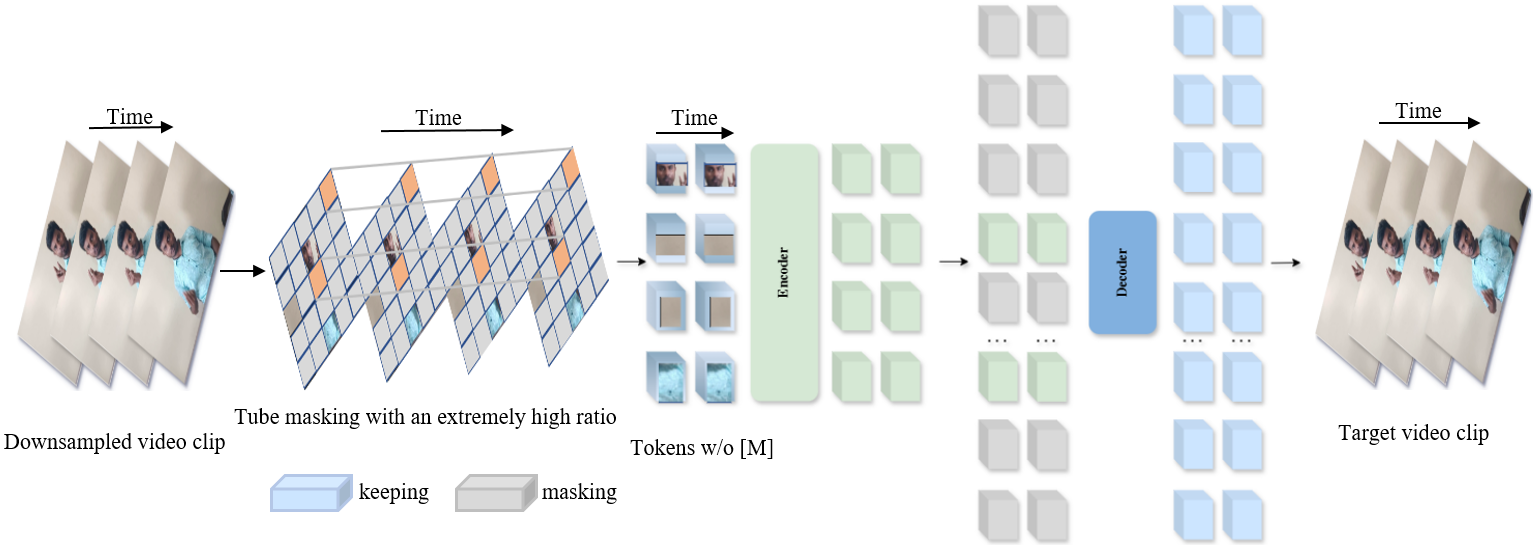} % Adjust both width and height
    \caption{VideoMAE Architecture for BdSLW60 Dataset Recognition \cite{tong2022videomaemaskedautoencodersdataefficient}}
    \label{fig:videoMAE}
\end{figure}

The idea of VideoMAE is found in Image-MAE~\cite{he2021maskedautoencodersscalablevision}, where the image masking strategy is described for better accuracy gain in recognition tasks. VideoMAE~\cite{huang2023mgmae} is a simple masked video autoencoder with an asymmetric encoder-decoder design. In order to handle the supplied sampled frames efficiently, it adds an additional cube embedding, which is visible in Figure~\ref{fig:videoMAE}.

Tube masking is implemented to address video redundancy, employing a high masking ratio (90-95\%) to avert information loss and improve reconstruction, especially in low-motion segments. The model incorporates an encoder that exclusively handles unmasked cubes and a streamlined decoder. Video segments are subjected to cube embedding, with merely 5-10\% of tokens inputted into the encoder. The model subsequently forecasts the masked tokens by reducing the discrepancy between target and projected clips. Tube masking surpasses alternative techniques by employing a uniform mask across the frames. The disordered tokens are reconstituted, and absent tokens are acquired through backpropagation. A compact decoder reconstructs video segments to assess performance. By encoding fewer tokens and using joint space-time attention~\cite{arnab2021vivitvideovisiontransformer} with a ViT backbone~\cite{dosovitskiy2021imageworth16x16words}, this method shortens the time needed for training.

%---------ViViT start-----
\subsubsection{Video Vision Transformer—ViViT architecture}
\begin{figure}[ht]
    \centering
    \includegraphics[width=1.0\columnwidth, height=0.3\textheight]{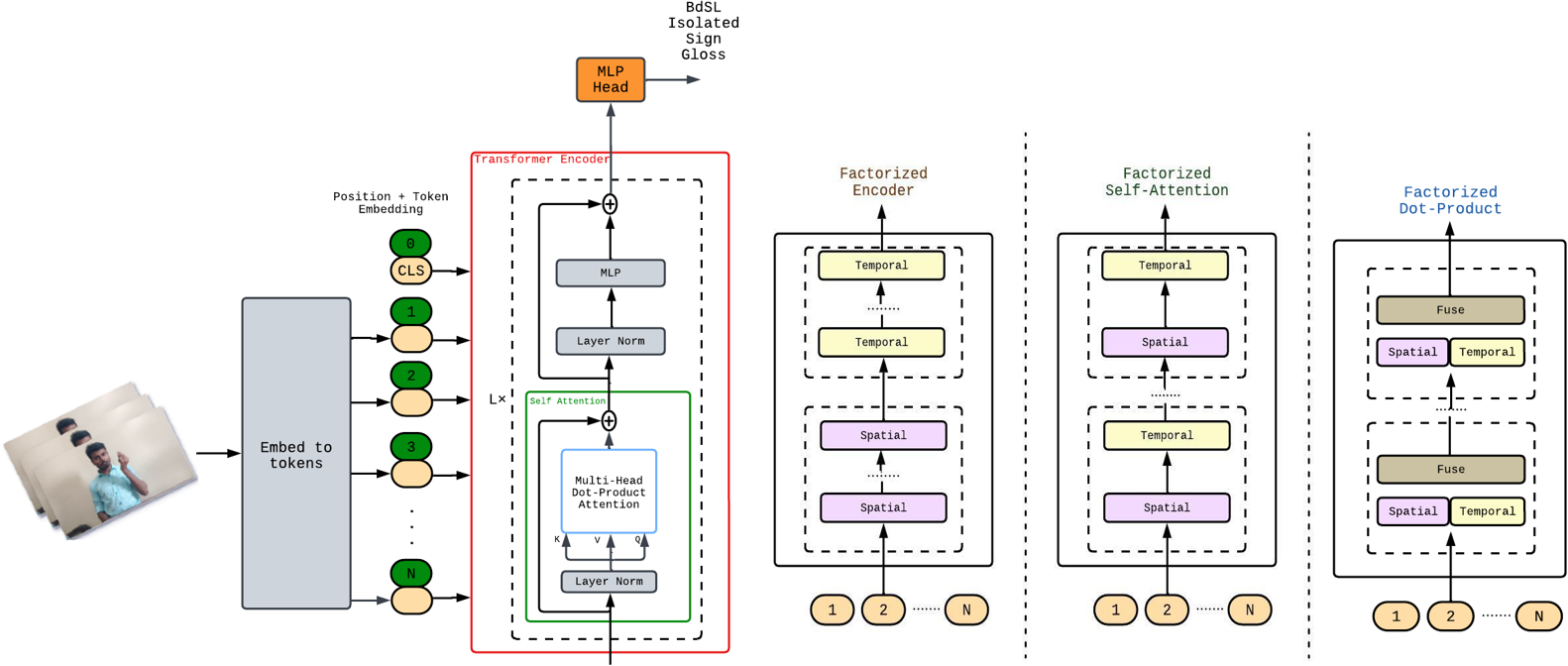} % Adjust both width and height
    \caption{Structure of the ViViT Model and Its Variations for BdSLW60 Recognition \cite{arnab2021vivitvideovisiontransformer}}
    \label{fig:vivit}
\end{figure}

Researchers have extended ViT~\cite{dosovitskiy2021imageworth16x16words}, originally developed for image classification, to create transformer-based models for video classification, as shown in Figure~\ref{fig:vivit}. These models use self-attention in the encoder to capture long-range contextual relationships within video sequences. Earlier approaches tackled this challenge using deep 3D CNNs~\cite{8099985,Feichtenhofer_2020_CVPR} and by incorporating self-attention in later layers~\cite{girdhar2019videoactiontransformernetwork, wang2018nonlocalneuralnetworks, wu2019longtermfeaturebanksdetailed}.

ViViT improves the Vision Transformer by integrating attention variations specifically designed for video data. It employs solely the encoder of the transformer~\cite{vaswani2023attentionneed}, analysing video clips $\mathcal{V} \in \mathbb{R}^{T \times H \times W \times C}$ transformed into token sequences $\mathcal{\tilde{Z}} \in \mathbb{R}^{n_t \times n_h \times n_w \times d}$. Two techniques—uniform frame sampling and tubelet embedding—convert videos into non-overlapping tokens, with tubelet embedding more efficiently integrating the temporal dimension. A tubelet possesses dimensions \( t \times h \times w \), with the quantity of tubelets along each axis defined as \( n_t = \left\lfloor \frac{T}{t} \right\rfloor \), \( n_h = \left\lfloor \frac{H}{h} \right\rfloor \), and \( n_w = \left\lfloor \frac{W}{w} \right\rfloor \). Following the incorporation of positional embeddings, tokens are input into the encoder, where self-attention scores are calculated. The ViViT model comprises 12 encoder layers and 12 attention heads. Four attention mechanisms are examined:
\begin{enumerate}
\item Spatio-temporal Attention: Employs Multi-Head Self-Attention (MSA)~\cite{vaswani2023attentionneed} across all tokens, leading to quadratic complexity.
\item Factorised Encoder: Distinguishes between spatial and temporal tubelet processing, reducing floating-point operations while preserving global context. 
\item Factorised Self-Attention: Executes attention initially in the spatial domain, followed by the temporal domain, preserving Model 2’s complexity while enhancing parameter efficiency. 
\item Factorised Dot-Product Attention: Distributes attention heads evenly over spatial and temporal domains, optimising complexity and parameter quantity. 
\end{enumerate}
Ultimately, a multilayer perceptron (MLP) forecasts class labels during the training process. This method improves efficiency while preserving robust performance in video representation learning.
%------------timesformer------------------

\subsubsection{TimeSformer architecture}

\begin{figure}[h!]
    \centering
    \includegraphics[width=0.8\columnwidth, height=0.4\textheight]{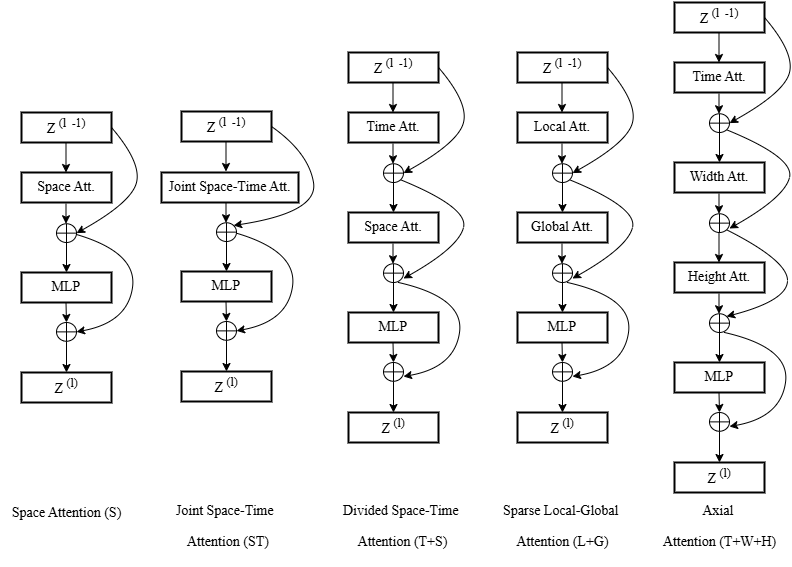}
    \caption{Five Self-Attention Blocks of TimeSformer \cite{bertasius2021spacetimeattentionneedvideo}}
    \label{fig:timesformer}
\end{figure} 
TimeSformer is a video categorization model that operates without convolution, utilizing a Vision Transformer. It separates video frames into N separate patches, uses learnable positional encoding, and a 12-layer transformer encoder to process them. The initial token, $\mathcal{Z}^{(0)}_{(0, 0)}$, functions as a classification token, with patch embeddings articulated as:
\begin{equation}
\mathbf{z}^{(0)}_{(p, t)} = \mathbf{E} \mathbf{x}_{(p, t)} + \mathbf{e}^{\text{pos}}_{(p, t)}
\label{eq:transformer_input}
\end{equation}
Self-attention improves computational efficiency by employing various attention techniques on patches, as demonstrated in Figure~\ref{fig:timesformer}. Spatial attention functions independently on a frame-by-frame basis, executing \( N + 1 \) query-key comparisons. Joint space-time attention encompasses both spatial and temporal dimensions; nonetheless, it is computationally demanding, necessitating \( N F + 1 \) comparisons, \( F \) denoting the total number of frames. Conversely, divided space-time attention successively analyzes temporal and spatial dimensions, attaining maximal accuracy with \( N + F + 2 \) comparisons. To improve efficiency, sparse local-global and axial attention equilibrate local and global emphasis while allocating attention across temporal, spatial width, and height dimensions. A multilayer perceptron with residual connections ultimately enhances the attention outputs.

\subsection{Fine-Tuning of video transformers}
Fine-tuning is a process that adapts a pretrained classification model to a new task by retraining it on task-specific data~\cite{8382272}. In this study, we fine-tuned video transformer models on the BdSLW60 dataset to classify isolated Bangladeshi Sign Language (BdSL) words. Leveraging the Hugging Face Transformers library, we initialized models with pretrained weights, modified the classification head to match the number of target classes, and utilized previously learned features to enhance generalization.

To further improve performance, we integrated a task-specific classification head into the final layer and systematically tuned key hyperparameters, including batch size, learning rate, and weight decay. A dynamic learning rate scheduler was employed to adjust learning rates throughout training, while model weights were optimized via backpropagation. To mitigate overfitting, an early stopping mechanism was implemented, enabling the training process to terminate when performance plateaued on the validation set. The full configuration of training parameters is summarized in Table~\ref{tab:hyperparameters}.
\begin{table}[ht]
    \centering
    \small
    \renewcommand{\arraystretch}{1.2} % Adjust row spacing for readability
    \setlength{\tabcolsep}{8pt} % Adjust column padding
    \begin{tabular}{ll}
        \toprule
        \textbf{Hyperparameter} & \textbf{Value} \\
        \midrule
        Training batch size & 2 \\
        Evaluation batch size & 2 \\
        Gradient accumulation steps & 4 \\
        Total effective batch size & 8 \\
        Initial learning rate & 5e-5 \\
        Weight decay & 0.01 \\
        Learning scheduler type & Linear \\
        Warm-up ratio & 0.1 \\
        Optimizer & AdamW \\
        Loss Function & Cross entropy loss \\
        \bottomrule
    \end{tabular}
    \vspace{5mm}
    \caption{Training Hyperparameters}
    \label{tab:hyperparameters}
\end{table}

\subsection{Model Evaluation}
Evaluating the performance of ML and DL models is crucial for both model development and deployment in real-world applications. Key evaluation metrics include accuracy, precision, recall, and the F1 score, each providing unique insights into model behaviour. Precision quantifies the proportion of correctly identified positive predictions, while accuracy measures the overall correctness across all classes. Recall evaluates the model’s ability to detect true positive instances, and the F1 score provides a harmonic mean of precision and recall, offering a balanced measure of classification performance~\cite{alnabih2024arabic}.

To comprehensively assess the fine-tuned transformer models, we computed accuracy, precision, recall, and F1 scores on the test datasets. In addition, confusion matrices were generated to visualize class-wise performance and identify potential misclassifications. The loss curve was also analyzed throughout training to monitor convergence behavior and detect signs of overfitting.

\section{Result Analysis}
The primary objective of this study was to identify the most effective pretrained video transformer model for classifying isolated BdSL signs using the BdSLW60 dataset, along with additional benchmark datasets. Notably, this work also presents the first benchmarking results for the BdSLW401 dataset, contributing a valuable reference point for future research in BdSL recognition. To evaluate the impact of data augmentation on model performance, we examined two video preprocessing strategies: one incorporating augmentation techniques such as random horizontal flipping and cropping, and another that excluded such modifications.

\begin{table}[ht]
    \centering
    \small
    \renewcommand{\arraystretch}{1.2} % Improve row spacing
    \setlength{\tabcolsep}{8pt} % Column padding
    \begin{tabular}{lccc}
        \toprule
        \textbf{Dataset} & \textbf{Train} & \textbf{Test} & \textbf{Val} \\
        \midrule
        BdSLW60 \cite{rubaiyeat2025bdslw60} & 7431 & 1276 & 600 \\
        BdSLW60 (10-fold) \cite{rubaiyeat2025bdslw60} & 8031 & 1276 & - \\
        BdSLW401 Front \cite{rubaiyeat2025bdslw401transformerbasedwordlevelbangla} & 38876 & 7833 & 4389 \\
        WLASL100 \cite{li2020wordleveldeepsignlanguage} & 1442 & 258 & 338 \\
        WLASL2000 \cite{li2020wordleveldeepsignlanguage} & 14289 & 2878 & 3916 \\
        LSA64 \cite{ronchetti2023lsa64argentiniansignlanguage} & 2304 & 640 & 256 \\
        \bottomrule
    \end{tabular}
    \vspace{5mm}
    \caption{Dataset splitting configurations.}
    \label{tab:dataset_splits}
\end{table}

The training hyperparameters were kept consistent across all experiments to ensure fair comparison. Table~\ref{tab:dataset_splits} summarizes the dataset partitioning strategies employed for training, validation, and testing. To support subject-independent evaluation, we implemented user-specific splits for BdSLW60, designating users U4 and U8 for testing and U5 for validation. For the LSA64 dataset, signer IDs 001 and 002 were assigned to the test set, while 10\% of the remaining samples were used for validation.

In the case of BdSLW401, we utilized the front-view subset and adhered to its original train/validation/test split, where users S04 and S08 were reserved for testing. For WLASL100 and WLASL2000, we adopted the official JSON-based partitions. Additionally, to assess model robustness, we conducted experiments using a 10-fold stratified version of the BdSLW60 dataset.

\subsection{Training approaches}
Training was conducted using computational resources comprising a 32GB GPU and 164GB of CPU memory, enabling efficient processing across both large- and small-scale datasets. Models trained with data augmentation consistently outperformed those trained without it. Validation results using VideoMAE and ViViT (Table~\ref{tab:model_test_accuracy}) confirmed that augmentation led to improved test accuracy, surpassing prior benchmarks such as those in \cite{rubaiyeat2025bdslw60}, which employed SVM and attention-based Bi-LSTM architectures.
Despite BdSLW60 offering more samples per gloss, its inherent class imbalance negatively affected model accuracy. This issue was partially addressed through stratified K-fold cross-validation, which preserved the original class distribution across all folds, thereby enhancing evaluation robustness.
Additionally, the use of 16-level relative quantization (RQ) in VideoMAE resulted in reduced accuracy, indicating a potential limitation of quantization-based compression in this context. 

\begin{table}[h!]
    \centering
    \small % Reduce the font size
    \renewcommand{\arraystretch}{1.2} % Adjust row height for better spacing
    \setlength{\tabcolsep}{3pt} % Adjust column padding for clarity
    \begin{tabular}{lcc|c|c}
        \toprule
        \multirow{2}{*}{\textbf{Model}} & \multirow{2}{*}{\textbf{Epoch}} & \multicolumn{2}{c}{\textbf{Test Accuracy}} \\
        \cmidrule(lr){3-4}
        & & \textbf{Aug: Yes} & \textbf{Aug: No} \\
        \midrule
        “MCG-NJU/videomae-base” & 20 & 84.95\% & 69.59\% \\
        “MCG-NJU/videomae-base-finetuned-kinetics” & 20 & \textbf{92.55\%} & 91.54\% \\
        “google/vivit-b-16x2-kinetics400” & 20 & 78.13\% & 74.37\% \\
        “MCG-NJU/videomae-base” with RQ & 20 & 82.05\% & 70.61\% \\
        \bottomrule
    \end{tabular}
    \vspace{5mm}
    \caption{Performance of VideoMAE and ViViT with and without augmentation.}
    \label{tab:model_test_accuracy}
\end{table}

\subsection{Comparing results among existing datasets} 
To enhance the validation of model performance, we extended our evaluation to include two public benchmark datasets: LSA64~\cite{ronchetti2023lsa64argentiniansignlanguage} and WLASL~\cite{li2020wordleveldeepsignlanguage}. Our primary focus, however, remained on isolated Bangla Sign Language recognition, using two datasets: BdSLW60 and BdSLW401. On the smaller-scale BdSLW60 dataset, our transformer-based approach achieved an accuracy of 95.5\%, surpassing existing baselines and highlighting the model’s strong capability in recognizing isolated signs.

On the more extensive BdSLW401 dataset, which comprises 401 Bangla sign words, the model attained an accuracy of 81.04\% after just 20 training epochs using the “MCG-NJU/videomae-base-finetuned-kinetics” architecture. This result is particularly promising given the dataset’s complexity and the prolonged training time of over five days.

To further benchmark the generalization capacity of our approach, we evaluated performance on LSA64, WLASL100, and WLASL2000. Our model demonstrated improvements over previous deep learning methods on both LSA64 and WLASL100. However, performance on WLASL2000 was comparatively limited, likely due to the dataset’s lower video quality and sparse sample distribution.

\begin{figure}
    \centering
    \includegraphics [width=0.8\columnwidth, height=0.4\textheight]{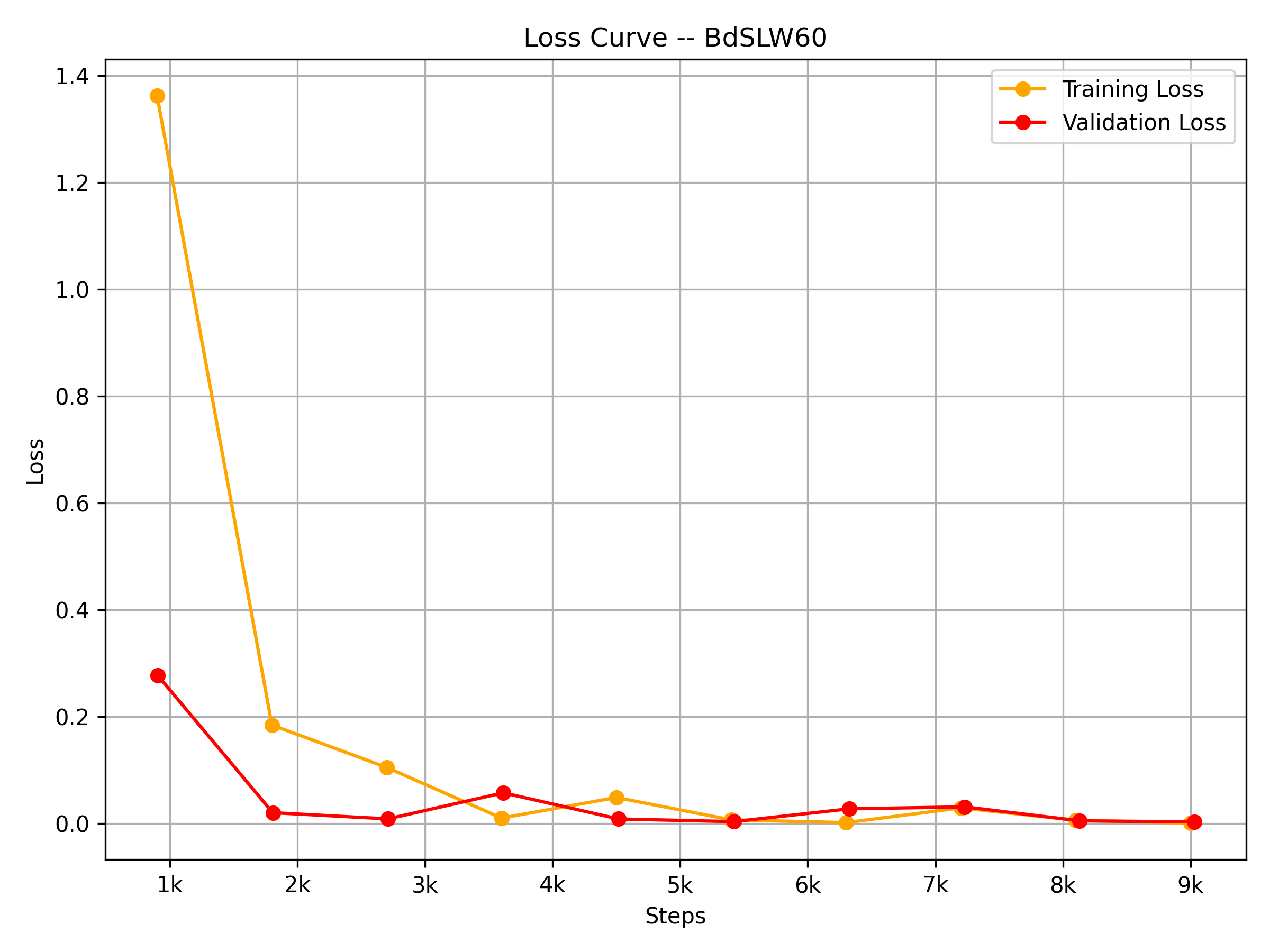}
    \caption{Loss Curve for Fold 9 of the BdSLW60 Dataset}
    \label{fig:bdslw60_loss}
\end{figure}

\begin{figure}
    \centering
    \includegraphics [width=1.0\columnwidth, height=0.5\textheight]{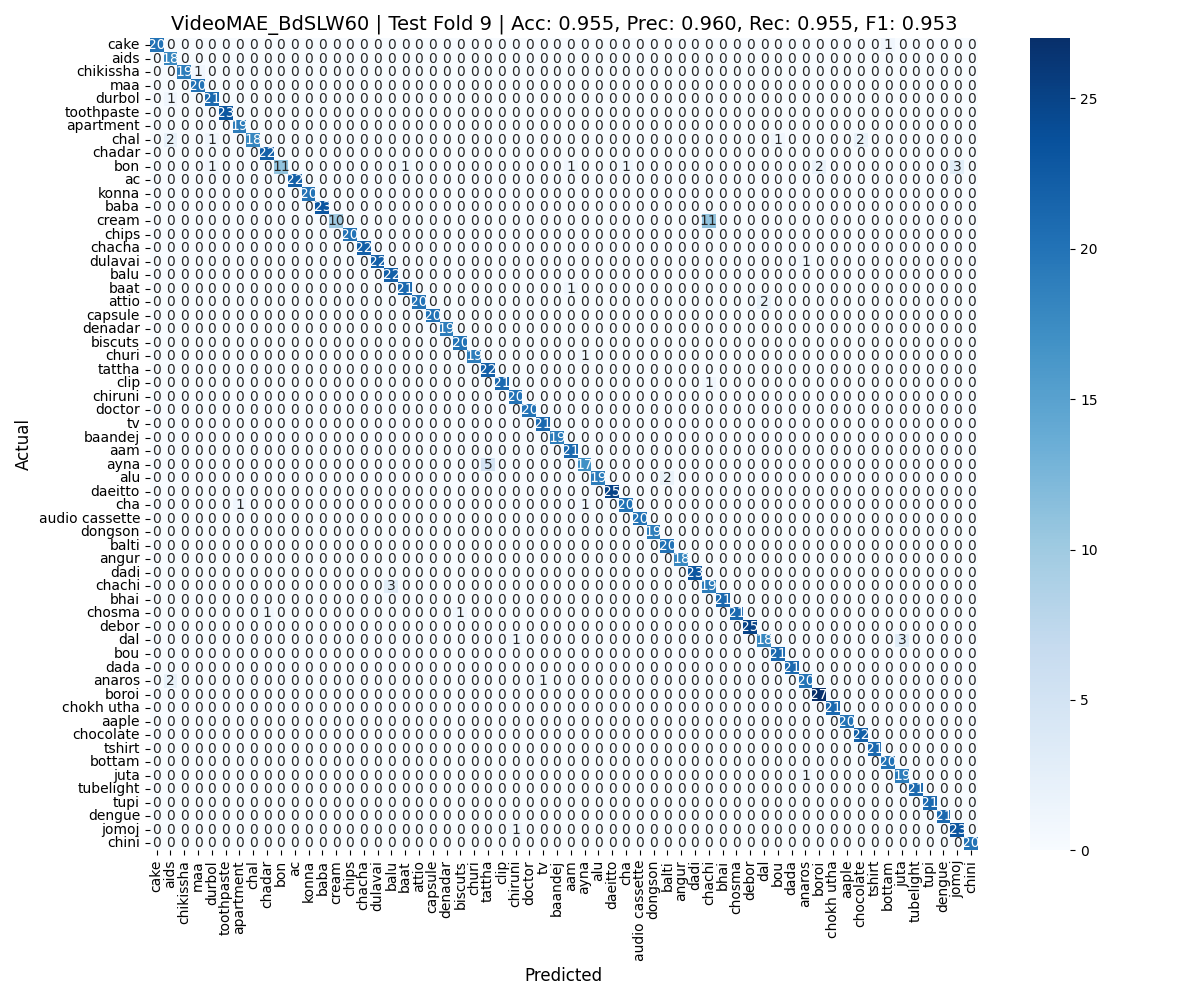}
    \caption{Confusion Matrix for Fold 9 of the BdSLW60 Test Set}
    \label{fig:BdSLW60_test_fold_9}
\end{figure}

Table~\ref{tab:results} summarizes the performance across all datasets, with boldface used to emphasize comparative accuracies. The loss curve and confusion matrices offer additional insights into training behavior and classification performance. On BdSLW60, the loss curve (Figure~\ref{fig:bdslw60_loss}) shows convergence and stability after approximately eight epochs, indicating strong generalization and minimal overfitting. The confusion matrix (Figure~\ref{fig:BdSLW60_test_fold_9}) further confirms accurate class-wise predictions.

\begin{figure}
    \centering
    \includegraphics[width=1.0\columnwidth, height=0.5\textheight]{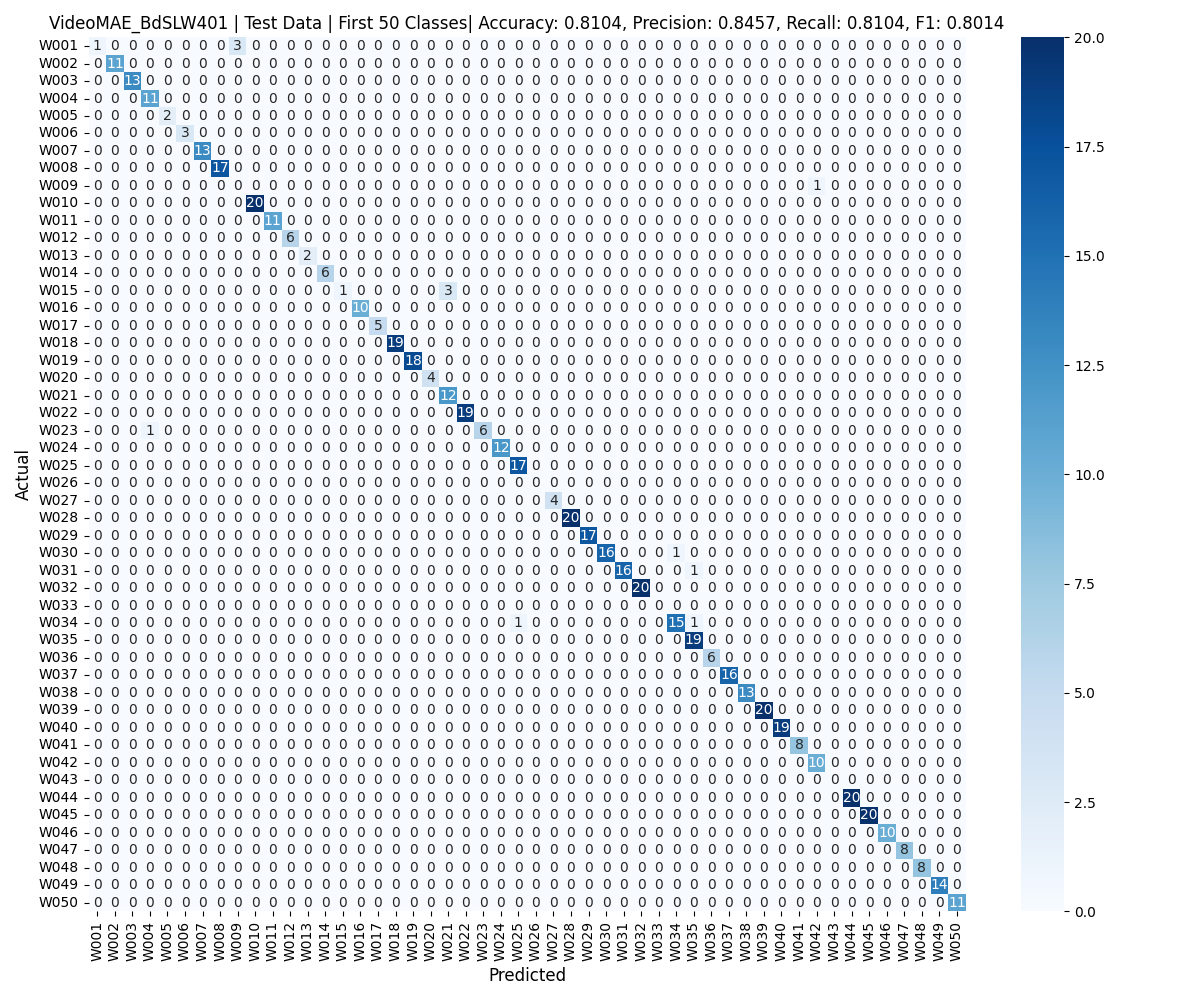}
    \caption{Confusion Matrix for BdSLW401 Test Set (Visualizing First 50 Classes)}
    \label{fig:bdslw401}
\end{figure}

Figure \ref{fig:bdslw401} illustrates the confusion matrix for BdSLW401, reflecting the model’s capability to scale effectively to a larger vocabulary. Together, these comparisons underscore the robustness and adaptability of transformer-based video models across both low- and high-resource sign language datasets.

%-------------- model comparisons
\begin{table*}[ht]
    \centering
    \small  % Reduced font size for better fit
    \renewcommand{\arraystretch}{1.4}  % Keep original row height
    \setlength{\tabcolsep}{5pt}  % Keep original column spacing
    \resizebox{\textwidth}{!}{  % Resize the table to fit within the text width
    \begin{tabular}{|l|l|l|c|c|c|c|}
        \hline
        \multirow{2}{*}{Papers} & \multirow{2}{*}{Dataset} & \multirow{2}{*}{Model} & \multicolumn{4}{c|}{Test Metrics} \\
        \cline{4-7}
        & & & Acc & Pre & Rec & F1 \\
        \hline
        %-------------------------
        \multirow{16}{*}{Our Work} & \multirow{4}{*}{BdSLW60 } & ``MCG-NJU/videomae-base'' & 93.6\% & 94.3\% & 93.7\% & 93.6\% \\
        & & ``MCG-NJU/videomae-base-finetuned-kinetics'' & \textbf{95.5\%} & 96.0\% & 95.5\% & 95.3\% \\
        & & ``google/vivit-b-16x2-kinetics400'' & 81.0\% & 84.9\% & 81.0\% & 80.7\% \\
        & & ``facebook/timesformer-base-finetuned-k400'' & 82.1\% & 86.2\% & 82.1\% & 81.9\% \\
        %-----------bdslw401
        \cline{2-7} % for drawing straight line
        & BdSLW401  & ``MCG-NJU/videomae-base-finetuned-kinetics'' & \textbf{81.04\%} & 84.57\% & 81.14\% & 80.14\% \\
        \cline{2-7} 
        % lsa64
        \multirow{4}{*}{} & \multirow{4}{*}{LSA64 } & ``MCG-NJU/videomae-base'' & 96.25\% & 96.8\% & 96.3\% & 96.1\% \\
        & & ``MCG-NJU/videomae-base-finetuned-kinetics'' & 97.65\% & 98.3\% & 97.6\% & 97.5\% \\
        & & ``google/vivit-b-16x2-kinetics400'' & 97.18\% & 97.7\% & 97.2\% & 97.1\% \\
        & & ``facebook/timesformer-base-finetuned-k400'' & \textbf{99.06\%} & 99.2\% & 99.1\% & 99.1\% \\
        
        %top 1,5,10
        \cline{2-7} % starting row
        \multirow{1}{*} & \multirow{1}{*} &  & Top 1 & Top 5 & Top 10 &\\
        \cline{4-6}

        %wlasl100
        \multirow{4}{*}{} & \multirow{4}{*}{WLASL 100 } & ``MCG-NJU/videomae-base'' & 51.55\% & 78.68\% & 86.05\% &  \\
        & & ``MCG-NJU/videomae-base-finetuned-kinetics'' & \textbf{66.7\%} & 88.0\% & 93.8\% &  \\
        & & ``google/vivit-b-16x2-kinetics400'' & 57.36\% & 86.05\% & 92.25\% & \\
        & & ``facebook/timesformer-base-finetuned-k400'' & 56.97\% & 84.5\% & 90.31\% &  \\
        \cline{2-7}
        % wlasl 2000
        \multirow{3}{*}{} & \multirow{3}{*}{WLASL 2000 } & ``MCG-NJU/videomae-base'' & 2.88\%  & 7.4\% & 9.6\% &  \\
        & & ``MCG-NJU/videomae-base-finetuned-kinetics'' & \textbf{6.9\%} & 12.4\% & 14.5\% &  \\
        & & ``google/vivit-b-16x2-kinetics400'' & 4.7\% & 10.4\% & 12.7\% &  \\
        
        \hline
        %--------------------------------------------

        % previous paper info: 
        \multirow{2}{*}{BdSLW60 \cite{rubaiyeat2025bdslw60}} & \multirow{2}{*}{BdSLW60} 
        & SVM & 67.6\% &  &  &  \\
        & & Attention-based bi-LSTM & \textbf{75.1\%} &  &  &  \\
        \hline
        % lsa64 base
        \multirow{1}{*}{LSA64 \cite{ronchetti2023lsa64argentiniansignlanguage}} & \multirow{1}{*}{LSA64} & ``HMM-GMM'' & 95.95\% &  &  &  \\

        % lsa64 3DGCN
        \multirow{1}{*}{3DGCN \cite{s22124558}} & \multirow{1}{*}{LSA64} & ``3D Graph Convolutional Neural Network'' & 94.84\% &  &  &  \\

        % lsa64 HWGAT
        \multirow{1}{*}{HWGAT \cite{patra2024hierarchicalwindowedgraphattention}} & \multirow{1}{*}{LSA64} & ``Hierarchical Windowed Graph Attention Network'' & \textbf{98.59\%} &  &  &  \\

        \hline
        %wlasl 100
        % top 1, 5, 10
        \multirow{1}{*} & \multirow{1}{*} &  & Top 1 & Top 5 & Top 10 &\\
        \cline{4-6}
        \multirow{8}{*}
        {WLASL \cite{li2020wordleveldeepsignlanguage}} & \multirow{4}{*}{WLASL 100} 
        & Pose-GRU & 46.51\% & 76.74\% & 85.66\% &  \\
        & & Pose-TGCN & 55.43\% & 78.68\% & 87.60\% &  \\
        & & VGG-GRU & 25.97\% & 55.04\% & 63.95\% & \\
        & & I3D & \textbf{65.89\%} & 84.11\% & 89.92\% &  \\
       % wlasl 2000
        \cline{2-6}
        & \multirow{4}{*}{WLASL 2000} 
        & Pose-GRU & 22.54\% & 49.81\% & 61.38\% &  \\
        & & Pose-TGCN & 23.65\% & 51.75\% & 62.24\% &  \\
        & & VGG-GRU & 8.44\% & 23.58\% & 32.58\% &  \\
        & & I3D & \textbf{32.48\%} & 57.31\% & 66.31\% &  \\
        \hline
    \end{tabular}
    }
    \caption{Experimental results on different models and datasets}
    \label{tab:results}
\end{table*}

%bdslw60 loss curve

%------------ confusion matrix bdslw60------

% bdslw401

\newpage
%----ablation study-------------
\subsection{Ablation Studies}
To optimize computational efficiency on our system, a batch size of two was employed. Larger batch sizes were avoided due to the associated increase in memory and storage requirements. During training, the entire pretrained model was initialized, and all layers were fine-tuned on our task-specific datasets. A learning rate scheduler dynamically adjusted the learning rate post-initialization, while the AdamW optimizer—with decoupled weight decay—was used to mitigate overfitting, making it particularly effective for large-scale model fine-tuning despite substantial memory demands.

We applied Frame Rate Correction (FRC) to BdSLW60 clips originally recorded at 15 and 24 FPS, converting them to 30 FPS for consistency. However, this led to frame duplication, which negatively affected generalization by distorting attention patterns and gradient updates. As a result, VideoMAE showed reduced accuracy on FRC clips compared to uncorrected samples. To address this, we introduced variability into the duplicated frames through random flipping, cropping, and scaling during preprocessing. This enhancement consistently improved performance, as shown in Table~\ref{tab:model_test_accuracy}.

Frame sampling rate played a crucial role in ensuring reliable and consistent input across clips, as detailed in Table~\ref{tab:Dataset details with fps}. For instance, WLASL2000 was trained for 200 epochs, with most outputs stabilizing after 40–50\% of the training cycle.

%--------------tab 6
\begin{table}[h!]
\centering
\small
\setlength{\tabcolsep}{7.0pt} % Reduce column spacing
\renewcommand{\arraystretch}{1.0} % Reduce row height
\vspace{5mm}  % Add vertical space between the table and the caption
\begin{tabular}{|l|c|l|c|c|}
\hline
\textbf{Dataset} & \textbf{FPS} & \textbf{Model} & \textbf{Sample rate (SR)} & \textbf{Clip Duration} \\ \hline
\multirow{3}{*}{BdSLW60}  & \multirow{3}{*}{30} & ViViT       & 4  & \multirow{3}{*}{4.27 s} \\ \cline{3-4}
                          &                     & VideoMAE    & 8  &                         \\ \cline{3-4}
                          &                     & Timesformer & 16 &                         \\ \hline
\multirow{3}{*}{BdSLW401} & \multirow{3}{*}{30} & ViViT       & 5  & \multirow{3}{*}{5.34 s} \\ \cline{3-4}
                          &                     & VideoMAE    & 10 &                         \\ \cline{3-4}
                          &                     & Timesformer & 20 &                         \\ \hline
\multirow{3}{*}{LSA64}    & \multirow{3}{*}{60} & ViViT       & 6  & \multirow{3}{*}{3.2 s}  \\ \cline{3-4}
                          &                     & VideoMAE    & 12 &                         \\ \cline{3-4}
                          &                     & Timesformer & 24 &                         \\ \hline
\multirow{3}{*}{WLASL}    & \multirow{3}{*}{25} & ViViT       & 4  & \multirow{3}{*}{5.2 s}  \\ \cline{3-4}
                          &                     & VideoMAE    & 8  &                         \\ \cline{3-4}
                          &                     & Timesformer & 16 &                         \\ \hline
\end{tabular}
\vspace{5mm}  % Optional: Add space below the table if needed
\caption{Dataset Details with FPS, Models, SR and Clip Durations}
\label{tab:Dataset details with fps}
\end{table}
Among the evaluated models, TimeSformer achieved a high accuracy of 99.06\% on the LSA64 dataset. Its superior performance is attributed to LSA64's balanced class distribution, 60 FPS recording rate, and clip lengths ranging between 90 and 180 frames. We standardized each clip to 3.2 seconds (192 frames), applying a sampling rate of 24 to extract 8 frames per video. Extending shorter clips to match this duration helped retain critical temporal information. In contrast, BdSLW60, recorded at 30 FPS with highly variable clip lengths (9–164 frames), posed additional challenges. Given that 99\% of BdSLW60 clips contain fewer than 128 frames, we extracted 4.27-second clips for uniformity.

On BdSLW60, TimeSformer’s performance was less competitive (82.1\% accuracy), likely due to its use of an 8-frame selection with a 16-frame sampling rate, which introduced redundancy. Similarly, ViViT suffered from diminished accuracy due to selecting 32 frames with a 4-sample rate, often leading to inefficient frame representation.

In contrast, VideoMAE demonstrated superior results by sampling 16 frames with an 8-frame rate, effectively reducing redundancy. Its use of a high-ratio masking strategy—retaining only 10\% of the tokens for encoding and reconstructing the remaining 90\% using mean squared error—allowed it to preserve essential temporal-spatial features while avoiding information overload.

The WLASL2000 dataset presented additional challenges, including class imbalance, small sample sizes, low-resolution clips, and irregular frame distributions, all of which contributed to degraded performance. These findings suggest that transformer-based video models perform optimally when trained on high-resolution datasets with evenly distributed frames and sufficient class representation.

\section{Conclusion and Future work}
%------------------------------
The fine-tuning process began with the BdSLW60 dataset, our primary data source. Raw videos were standardized to a common frame rate and segmented into shorter clips suitable for training. Augmentation techniques—such as random cropping, flipping, and scaling—were applied to improve model generalization.

Using BdSLW60, the MCG-NJU/videomae-base-finetuned-kinetics model achieved a test accuracy of 95.5\%, demonstrating strong performance in isolated BdSL recognition. To further evaluate model scalability, we introduce the first benchmark results on the BdSLW401 dataset, a larger and more diverse collection of 401 Bangla signs. On this dataset, the model achieved 81.04\% accuracy, with an F1 score of 80.14\%, recall of 84.57\%, and precision of 81.14\%.

Additional experiments on LSA64 and WLASL showed that factors such as frame distribution, video quality, sample size, and architecture significantly affect recognition accuracy. Overall, our approach outperforms prior methods in Bangla word-level sign language recognition. Future work will extend to sentence-level BdSL recognition and real-time translation applications.

\bibliographystyle{unsrt}  
\bibliography{references}  %%% Remove comment to use the external .bib file (using bibtex).
\end{document}